\documentclass[10pt,twocolumn,letterpaper]{article}

\usepackage{cvpr}
\usepackage{times}
\usepackage{epsfig}
\usepackage{graphicx}
\usepackage{amsmath}
\usepackage{amssymb}
\usepackage{mathtools}
\usepackage{times}
\usepackage{epsfig}
\usepackage{graphicx}
\usepackage{comment}
\usepackage{amsmath,amssymb} 
\usepackage{color}
\usepackage{amsmath}
\usepackage{amssymb}
\usepackage{xcolor}
\usepackage{subcaption}
\usepackage{dblfloatfix}
\usepackage{setspace}
\usepackage{bbm}
\usepackage{gensymb}
\usepackage{booktabs}
\usepackage{cite}

\usepackage[ruled,lined,linesnumbered]{algorithm2e}

\newcommand{\myparagraph}[1]{{\vspace{0.5em} \noindent \bf #1}}

\usepackage[breaklinks=true,bookmarks=false,colorlinks]{hyperref}

\cvprfinalcopy 

\def\cvprPaperID{****} 

\begin{document}
\title{2nd Place Solution for \\ Waymo Open Dataset Challenge - 2D Object Detection}

\author{Sijia Chen$^{*}$ \quad Yu Wang$^{*}$ \quad Li Huang \quad Runzhou Ge \\ \quad Yihan Hu  \quad Zhuangzhuang Ding \quad Jie Liao\\
[1.0ex]
Horizon Robotics Inc.\\
{\tt\small chensijia94@gmail.com \quad yuwangrpi@gmail.com}
}

\twocolumn[{
\maketitle
\begin{center}
    \centering 
    \vspace{-0.3in}
    \includegraphics[width=1\linewidth]{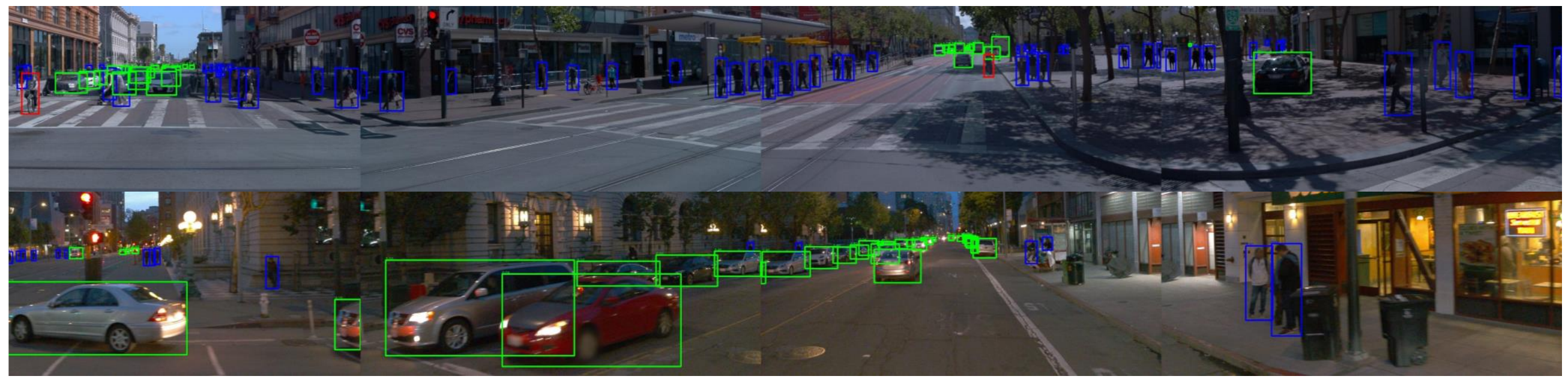}
    \vspace{-0.25in}
    \captionof{figure}{Detection results on the Waymo Open Dataset. Each row shows detections of $4$ cameras at the same timestamp. The green, blue, and red boxes denote vehicle, pedestrian, and cyclist classes, respectively.}
    \label{teaser}
\end{center}
}] 
\newcommand\blfootnote[1]{%
  \begingroup
  \renewcommand\thefootnote{}\footnote{#1}%
  \addtocounter{footnote}{-1}%
  \endgroup
}

\blfootnote{$\prescript{*}{}{}$These authors contributed equally to this work.}

\begin{abstract}
A practical autonomous driving system urges the need to reliably and accurately detect vehicles and persons. In this report, we introduce a state-of-the-art 2D object detection system for autonomous driving scenarios. Specifically, we integrate both popular two-stage detector and one-stage detector with anchor free fashion to yield a robust detection. Furthermore, we train multiple expert models and design a greedy version of the auto ensemble scheme that automatically merges detections from different models. Notably, our overall detection system achieves 70.28 L2 mAP on the Waymo Open Dataset v1.2, ranking the 2nd place in the 2D detection track of the Waymo Open Dataset Challenges.
\end{abstract}


\section{Introduction}
The Waymo Open Dataset challenges attracted many participants in the field of computer vision and autonomous driving. The Waymo Open Dataset~\cite{waymodata} that is used in the competition provides high-quality data collected by multiple LiDAR and camera sensors in real self-driving scenarios. In the 2D detection track, three classes: vehicle, pedestrian, and cyclist are annotated with tight-fitting 2D bounding boxes based on the camera images. In self-driving applications accurately and reliably detecting vehicles, cyclists and pedestrians is of paramount importance. Towards this aim, we develop a state-of-the-art 2D object detection system in this challenge.

\section{Our Solution}
\subsection{Base Detectors}
With the renaissance of deep learning based object detector, two mainstream frameworks, \ie, one-stage detector and two-stage detector, have dramatically improved both accuracy and efficiency. To fully leverage different detection frameworks, we employ the state-of-the-art two-stage detector Cascade R-CNN~\cite{cascade} and one-stage detector CenterNet~\cite{centernet} with anchor-free fashion. Cascade R-CNN employs a cascade structure for classification and box regression of proposed candidates, which is good at precisely localizing object instances. In contrast to Cascade R-CNN, CenterNet is anchor-free and treats objects as points with properties, which may be better suited for detecting small objects and objects in crowded scenes. We argue that these two different mechanisms have fair diversity on their detections so that the results can be complementary. We respectively produce detections using these two frameworks and then fuse them as the final detection results.

\subsubsection{Cascade R-CNN}
Cascade is a classical architecture that is demonstrated to be effective for various tasks. Among the object detection counterparts, Cascade R-CNN builds up a cascade head based on the Faster R-CNN~\cite{faster_rcnn} to refine detections progressively. Since the proposal boxes are refined by multiple box regression heads, Cascade R-CNN is skilled in precisely localizing object instances. In this challenge, we utilize the Cascade R-CNN as our two-stage detector counterpart considering its superiority.

\subsubsection{CenterNet}

\begin{figure}
\includegraphics[width=0.9\linewidth]{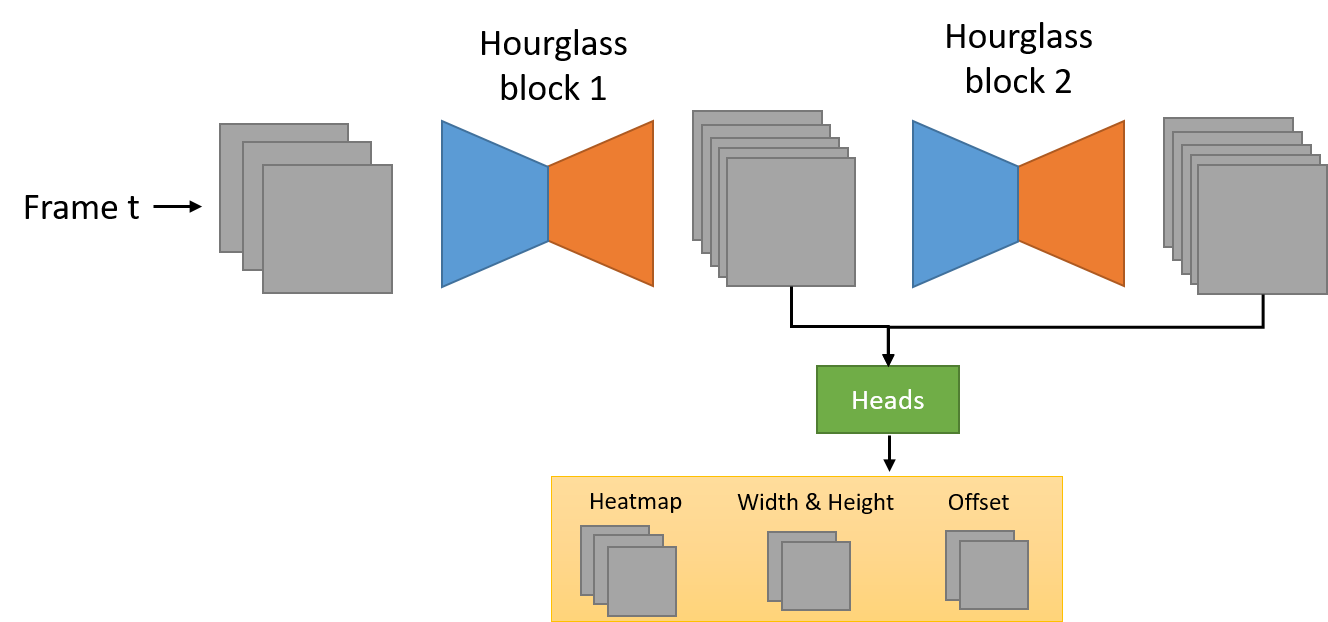}
\caption{CenterNet~\cite{centernet} detector with Hourglass-104 as the backbone. Two hourglass blocks are stacked and the first one only serves as providing auxiliary loss during training.}
\label{fig:centernet} 
\end{figure}

Recently, anchor-free object detectors become popular due to its simplicity and flexibility. CenterNet~\cite{centernet} detects objects via predicting the central location as well as spatial sizes of object instances. Since CenterNet does not need the Non-Maximum Suppression (NMS) as the post-processing step, it may be more suitable for crowded scenes where the NMS may wrongly suppress positive boxes if the threshold is not appropriately set. In this challenge, we employ CenterNet as our one-stage detector counterpart whose framework is shown in Figure~\ref{fig:centernet}. In contrast to the original CenterNet, we use the Gaussian kernel as in \cite{liu2019trainingtimefriendly} which takes into account the aspect ratio of the bounding box to encode training samples.

\subsection{Greedy Auto Ensemble}
\begin{figure}
\includegraphics[width=0.9\linewidth]{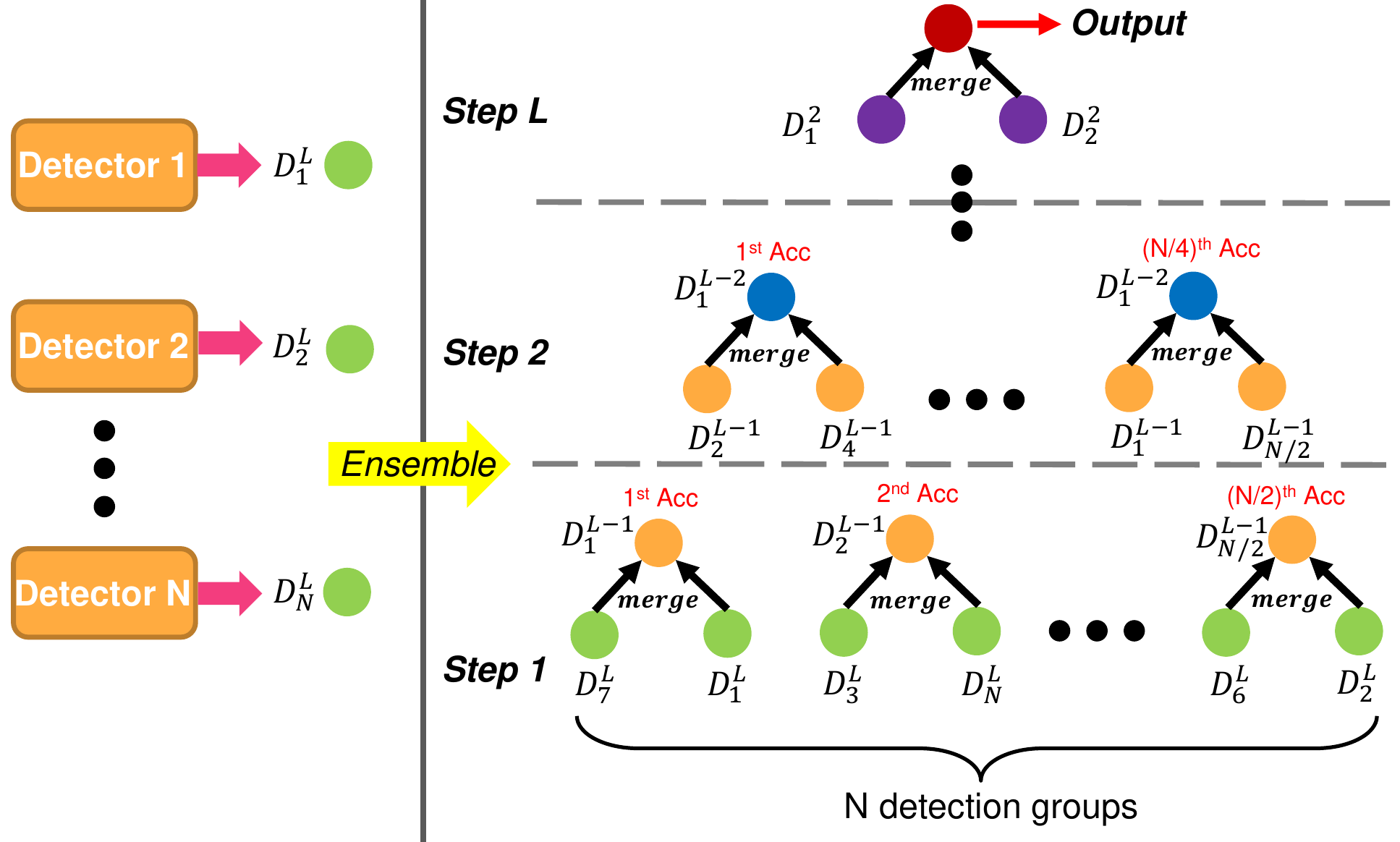}
\caption{Greedy Auto Ensemble. For $N$ groups of detections, we merge each two of them iteratively according to the resulting accuracy, until only one group of detections is left. This greedy scheme is more efficient compared with the original Auto Ensemble in \cite{openimg}.}
\label{fig:GAE} 
\end{figure}

We design a greedy version of the auto ensemble scheme~\cite{openimg} that automatically merges multiple groups of detections according to their detection accuracies, as shown in Figure~\ref{fig:GAE}. We note that a group of detections represents the detection results generated from a unique detector framework or a specific inference scheme (\eg, testing with specific image scales) of the same detector framework. As in~\cite{openimg}, we consider each group of detections as a node of a binary tree. Let $\mathcal{S}_{l}=\{\mathcal{D}^{k}_{l}\}_{k=1}^{N_{l}}$ be all the detection groups at the $l$-th level of the binary tree, where $\mathcal{D}^{k}_{l}$ is the $k$-th group of detections and $N_{l}$ is the number of detection groups at the $l$-th level. We denote $L$ as the number of levels of the binary tree. Note that $\mathcal{S}_{L}=\{\mathcal{D}^{k}_{L}\}_{k=1}^{N_{L}}$ indicates all the leaf nodes whose detection results are generated from different models and $N_{L}$ stands for the total number of detector frameworks and inference schemes used for the ensemble. For each node $\mathcal{D}^{k}_{l}$, it will be evaluated on the validation set and the corresponding accuracy is $A^{k}_{l}$ that is calculated based on the mAP metric. We iteratively merge every pair of children nodes into one parent node in each level of the binary tree until the root node is reached, where the root node serves as the final detection results. Different from~\cite{openimg}, our method determines the hierarchical relations of the binary tree dynamically and greedily and therefore reduces much search space. To be more specific, at the $l$-th level, we treat two nodes $\mathcal{D}^{i}_{l}$ and $\mathcal{D}^{j}_{l}$ as siblings if $\mathcal{D}^{i}_{l}$ and $\mathcal{D}^{j}_{l}$ are available in the candidate node set $\mathcal{C}_{l}$ and the $D^{k}_{l-1}=merge(\mathcal{D}^{i}_{l},\mathcal{D}^{j}_{l})$ yields the best accuracy so far, in which $D^{k}_{l-1}$ is the parent node of $\mathcal{D}^{i}_{l}$ and $\mathcal{D}^{j}_{l}$ and $merge(\cdot,\cdot)$ is the merge operation. After merging $\mathcal{D}^{i}_{l}$ and $\mathcal{D}^{j}_{l}$, we delete them from $\mathcal{C}_{l}$ and add $D^{k}_{l-1}$ into $\mathcal{C}_{l-1}$.

For the merge operation $merge(\cdot,\cdot)$, we search several candidate operations and employ the operation that yields the best accuracy. Given two nodes $\mathcal{D}^{i}_{l}$ and $\mathcal{D}^{j}_{l}$, we define $merge(\cdot,\cdot)$ as:

\begin{equation}
\label{eqn:eq_base}
\begin{aligned}
&merge(\cdot,\cdot)=\underset{o \in \mathcal{O}}{\operatorname{argmax}}\ mAP(o(\mathcal{D}^{i}_{l},\mathcal{D}^{j}_{l})), \\
&s.t. \ \ \ \mathcal{O}=\{nms, adj{\text -}nms, nmw{\text -}naive, o1, o2\},
\end{aligned}
\end{equation}
where $\mathcal{O}$ is the operation set used in our method. $nms$ and $adj{\text -}nms$ denotes the traditional NMS and the Adj-NMS~\cite{openimg} respectively. $nmw{\text -}naive$ is a simplified version of non-maximum weighted (NMW)~\cite{cad} that only use confidence scores as weights to merge multiple boxes into one box. In case the detection performance may be degraded after merging, we also introduce $o1(\mathcal{D}^{i}_{l},\mathcal{D}^{j}_{l})=\mathcal{D}^{i}_{l}$ and $o2(\mathcal{D}^{i}_{l},\mathcal{D}^{j}_{l})=\mathcal{D}^{j}_{l}$. The overall algorithm of the greedy auto ensemble is presented in Algorithm~\ref{alg:alg1}.

\begin{algorithm}
	\caption{Greedy Auto Ensemble}
	\label{alg:alg1}
	\setstretch{1.158}
	\SetAlgoLined
	\SetKwInOut{Input}{Input} \SetKwInOut{Output}{Output} 
    \Input{$\mathcal{S}_{L}=\{\mathcal{D}^{k}_{L}\}_{k=1}^{N_{L}}$: The detection results generated from $N_L$ different models.
    \ }
	\Output{$\mathcal{S}_{1}=\mathcal{D}^{1}_{1}$: The final detection results.}
	// \textbf{Initialization:} 
	// Candidate node set initialization\\
	$C_{L} \leftarrow \mathcal{S}_{L}$ \\
	\For{$l \leftarrow 1 \ to \ L-1$}
	{$C_{l} \leftarrow \emptyset$}
	\For{$l \leftarrow L \ to \ 2$} {
	\While{$\vert C_{l} \vert > 1$}{
	$\mathcal{D}^{i}_{l},\mathcal{D}^{j}_{l} \leftarrow \underset{\mathcal{D}^{i}_{l},\mathcal{D}^{j}_{l} \in C_{l}}{\operatorname{argmax}}\ mAP(merge(\mathcal{D}^{i}_{l},\mathcal{D}^{j}_{l}))$\\
	$ C_{l-1} \leftarrow C_{l-1} \cup \{merge(\mathcal{D}^{i}_{l},\mathcal{D}^{j}_{l})\}$\\
	$ C_{l} \leftarrow C_{l} \setminus \{\mathcal{D}^{i}_{l},\mathcal{D}^{j}_{l}\}$
	}
	\If{$\vert C_{l} \vert = 1$}{
	$ C_{l-1} \leftarrow C_{l-1} \cup C_{l}$
	}
	}
$S_1 \leftarrow C_{1}$\\
\textbf{Return:} $S_1$
\end{algorithm}

\subsection{Expert Model}
Data distribution is highly imbalanced in the Waymo Open Dataset \cite{waymodata}. For example, there are $1.7M$, and $6M$ instances for the vehicle and pedestrian classes but only $50K$ instances for the cyclist class in the training set. As a result, the cyclist class could be overwhelmed by pedestrian or vehicle samples in training, leading to poor performance on the cyclist class. To solve this problem, we train multiple expert models for the cyclist, pedestrian, and vehicle classes, respectively. Since the Waymo Open Dataset also provides context information for each image frame such as time of the day (\eg daytime and nighttime). We also train additional daytime and nighttime expert models using only daytime and nighttime training images, respectively.

\subsection{Anchor Selection}
In Cascade R-CNN, the anchors are predefined manually. By default, the aspect ratios are set to 0.5, 1, and 2. We add two more anchor aspect ratios of $0.25$ and $0.75$ for the vehicle expert model since we observe some vehicles with very elongated shapes. CenterNet is free of the anchor selection problem.

\begin{figure}
\includegraphics[width=1.0\linewidth]{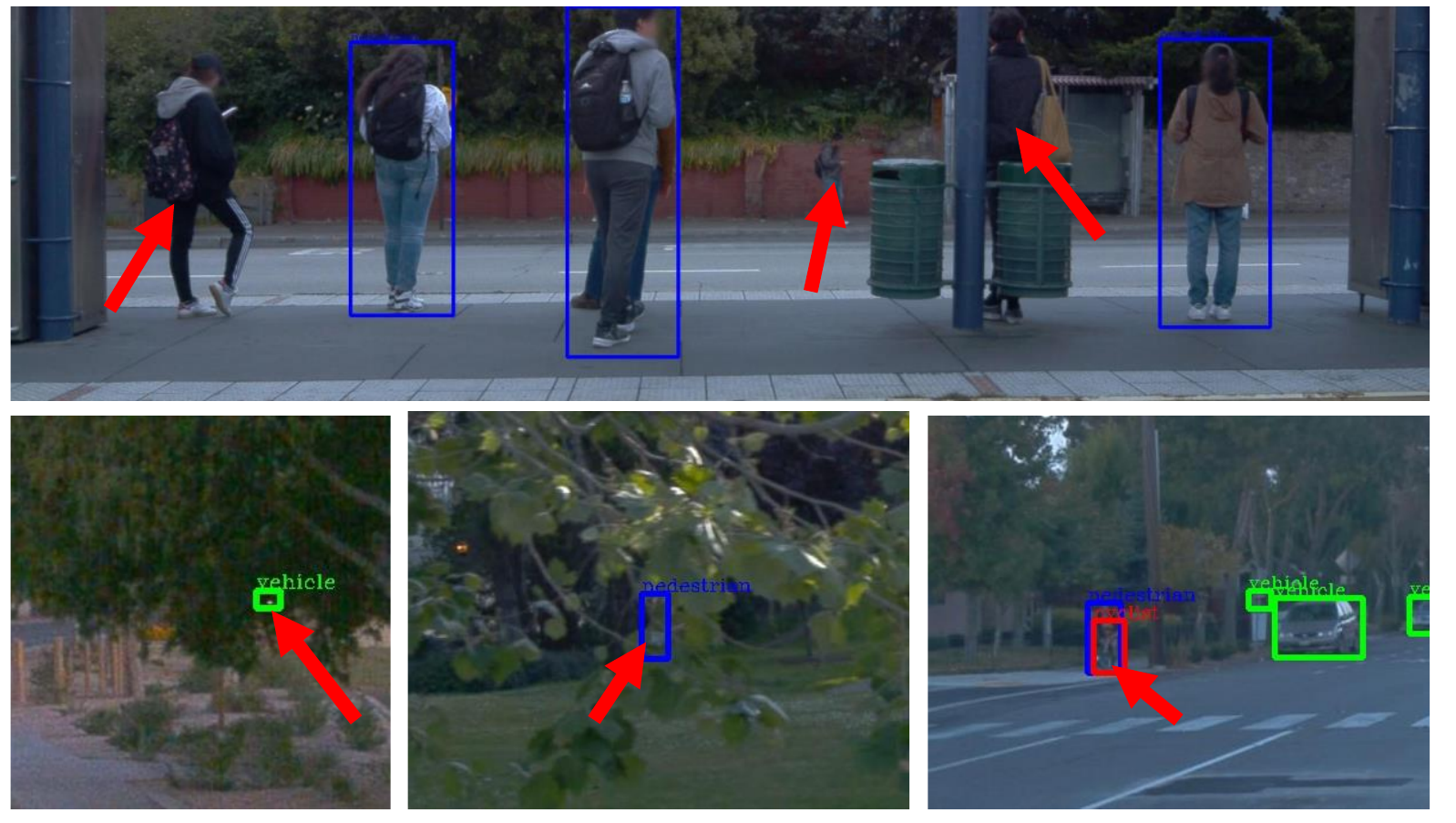}
\caption{Visualization of some hard examples or inaccurate annotations.}
\label{fig:fig2} 
\end{figure}

\subsection{Label Smoothing}
After visually inspecting the annotations, we noticed some hard examples and inaccurate or missing annotations as shown in Figure~\ref{fig:fig2}, which may cause problems for the training. Therefore, we employ label smoothing to handle this problem during training.

\section{Experiments}
\subsection{Dataset and Evaluation}
\myparagraph{Dataset.} The Waymo Open Dataset v1.2 \cite{waymodata} contains $798$, $202$ and $150$ video sequences in the training, validation, and testing sets, respectively. Each sequence has 5 views of side left, front left, front, front right, and side right, where each camera captures $171$-$200$ frames with the image resolution of $1920 \times 1280$ pixels or $1920 \times 886$ pixels. Our models are pre-trained on the COCO dataset then fine-tuned on the Waymo Open Dataset v1.2. Due to limited computational resources, we sample $1$ frame of every $10$ frames from the training set to form a \emph{mini-train} set, which is used to train the Cascade R-CNN and some CenterNet models. We also sample $1$ frame of every $20$ frames from the validation set to form a \emph{mini-val} set for ablation experiments. In our solution, temporal cues are not used.

\myparagraph{Evaluation Metrics.} According to the Waymo Open Dataset Challenge 2D Detection track, we report detection results on the Level 2 Average Precision (AP) that averages over vehicle, pedestrian, and cyclist classes. The positive IoU thresholds are set to $0.7$, $0.5$, and $0.5$ for evaluating vehicles, cyclists, and pedestrians, respectively.

\subsection{Implementation Details}
\label{subsec:implementation}
\myparagraph{Cascade R-CNN Detector.} For the Cascade R-CNN detector, we adopt the implementation of Hybrid Task Cascade \cite{htc} in mmdetection \cite{mmdet} with disabled semantic segmentation and instance segmentation branches, as pixel-wise annotations are not available in this challenge. We use the ResNeXt-$101$-$64\times4$d \cite{resnext} with deformable convolution \cite{dcn} as the backbone network. We train one main model with all three classes and three expert models for vehicle, pedestrian, and cyclist classes all on the \emph{mini-train} set, respectively. The main model is trained for $10$ epochs with a warm-up learning rate from $1.7e$-3 to $5e$-3, and the learning rate is then decayed by a factor of $0.1$ at the $7$th epoch and $9$th epoch, respectively. We train the expert models for 7 epochs with a learning rate warmed up from $3.3e$-5 to $1e$-4 and then decayed by a factor of $0.1$ at the $5$th epoch. We also use the multi-scale training, where the long dimension is resized to $1600$ pixels while the short dimension is randomly selected from $[600, 1000]$ pixels without changing the original aspect ratio. Label smoothing and random horizontal flipping are also applied in training. The batch size for all models is set to $8$.

During inference, we resize the long dimension of each image to $2400$ pixels and keep its original aspect ratio. We use multi-scale testing with $3$ scale factors of $0.8, 1.0, 1.2$ as well as the horizontal flipping for all models, except for the vehicle expert which only adopts the horizontal flipping. For each model, we first use the class-aware soft-NMS to filter out overlapped boxes. To merge detections generated by  different models, we employ the greedy auto ensemble for the pedestrian and cyclist classes, respectively, and use the Adj-NMS for the vehicle class.

\myparagraph{CenterNet Detector.} For the CenterNet detector, the image size is set to $768 \times 1152$ pixels during training, and the learning rate is set to $1.25e$-4. To save computational resources, we first train the CenterNet detector with COCO pretrained weights on the \emph{mini-train} set for 25 epochs and use it as the base model. We then fine-tune $3$ expert models based on the base model: nighttime expert, daytime expert, pedestrian+cyclist expert, respectively. We also fine-tune another $4$ expert models based on the base model using the validation set, the training set, the training set with only pedestrian and cyclist classes, and the training set with only nighttime images for 8-10 epochs. In inference, the horizontal flipping and multi-scale testing with scale factors of $0.5$, $0.75$, $1$, $1.25$, $1.5$ are used. To sum up, we train $8$ CenterNet models in total and merge their detections into one group of detections using the weighted boxes fusion (WBF)\cite{solovyev2019weighted}.

\myparagraph{Ensemble.} One-stage detector and two-stage detector each produces an independent group of detections. To merge the two groups of detections into the final result we use Adj-NMS for the vehicle and pedestrian classes, respectively, and utilize WBF for the cyclist class.

\subsection{Results}
To study the effect of each module used in our solution, we perform ablation experiments on the \emph{mini-val} set as shown in Table~\ref{tbl:ablation}. We first evaluate the Cascade R-CNN baseline with label smoothing that achieves $59.71$ AP/L2. We further improve performance from $59.71$ to $61.04$ on AP/L2 by utilizing the commonly used inference schemes of class-aware soft-NMS and multi-scale testing. To assess the greedy auto ensemble, we merge the results of baseline with those of expert models, which leads to a notable improvement of $2.24$ AP/L2. Finally, we fuse the detections of the Cascade R-CNN and CenterNet, which further improves $1.44$ AP/L2 compare to the results of CenterNet. It demonstrates the effectiveness of combining the one-stage and two-stage detectors.

The 2D detection track is quite competitive among all the five tracks in the Waymo Open Dataset Challenge.
To compare our final submitted result with other competitors, we show the leaderboard of the Waymo Open Dataset Challenge - 2D Detection Track in the Table~\ref{tbl:final}. It is seen that our overall detection system achieves superior detection results and ranks the 2nd place among all the competitors.

\begin{table}[t]
\begin{center}
\setlength\tabcolsep{6pt}
\begin{tabular}{l|c}
\hline
Method               &AP/L2 \\
\hline\hline
Cascade R-CNN baseline               &59.71\\
+ class-aware softnms      &60.42\\
+ multi-scale testing             &61.04\\
+ GAE + Expert Models     &63.28\\
\hline
CenterNet    &64.83\\
\hline
Our Solution          &66.27\\

\hline
\end{tabular}
\end{center}
\caption{Ablation study on the \emph{mini-val} set. The ``GAE'' stands for the greedy auto ensemble. The result of CenterNet is obtained by merging the detections of the $8$ models described in Section~\ref{subsec:implementation} using the WBF. Our solution indicates the compound of the above methods.} 
\label{tbl:ablation}
\end{table}

\begin{table}[t]
\begin{center}
\setlength\tabcolsep{10pt}
\begin{tabular}{l|c|c}
\hline
Method Name         &AP/L1          &AP/L2 \\
\hline\hline
RW-TSDet            &\textcolor{red}{\bf 79.42}          &\textcolor{red}{\bf 74.43}\\
HorizonDet (Ours)     &\textcolor{blue}{\bf 75.56}      &\textcolor{blue}{\bf 70.28}\\
SPNAS-Noah          &75.03          &69.43 \\
dereyly\_alex\_2    &74.61          &68.78 \\
dereyly\_alex       &74.09          &68.17 \\

\hline
\end{tabular}
\end{center}
\caption{Leaderboard of the Waymo Open Dataset Challenge - 2D Detection Track \cite{leaderboard}, where we only list the top-5 entries. The top-2 results are highlighted in red and blue colors, respectively.} 
\label{tbl:final}
\end{table}

\section{Conclusion}
In this report, we present a state-of-the-art 2D object detection system for autonomous driving scenarios. Specifically, we utilize both popular one-stage and two-stage detectors to yield robust detections of vehicles, cyclists and pedestrians. We also employ various ensemble approaches to merge detections from various models. Our overall detection system achieved the 2nd place in the 2D detection track of the Waymo Open Dataset Challenges.

\end{document}